\documentclass[11pt,a4paper]{article}
\usepackage[hyperref]{emnlp2020}
\usepackage{times}
\usepackage{latexsym}
\usepackage{framed}
\usepackage{enumitem}
\usepackage{subcaption}
\usepackage{graphicx}
\usepackage{booktabs}
\usepackage{multirow}
\usepackage{multicol}
\usepackage{enumitem}

\usepackage{xspace}
\usepackage{amsmath}
\usepackage{amssymb}
\usepackage{subcaption}

\newcommand{\fulllabel}[2]{\textbf{#1}\newline\textbf{#2}}

\usepackage{microtype}

\aclfinalcopy 
\def\aclpaperid{1540} 

\newcommand{\abdnli}{$\alpha$NLI\xspace}
\newcommand{\snliat}{SNLI$^{@3}$\xspace}
\newcommand{\mnlimat}{MNLI-m$^{@3}$\xspace}

\newcommand{\chaosNLIs}{ChaosNLI-S\xspace}
\newcommand{\chaosNLIm}{ChaosNLI-M\xspace}
\newcommand{\chaosNLIa}{ChaosNLI-$\alpha$\xspace}

\newcommand{\bq}{\mathbf{q}}
\newcommand{\bp}{\mathbf{p}}
\newcommand{\bm}{\mathbf{m}}
\newcommand{\kl}[1]{\textsc{kl}\left(#1\right)}
\newcommand{\jsd}[1]{\textsc{jsd}\left(#1\right)}

\title{What Can We Learn from Collective Human Opinions \\ on Natural Language Inference Data?}

\author{Yixin Nie \ \ \ \ \ \ Xiang Zhou \ \ \ \ \ \ Mohit Bansal \\
  Department of Computer Science \\
University of North Carolina at Chapel Hill \\
  \texttt{\{yixin1, xzh, mbansal\}@cs.unc.edu} \\
}

\date{}

\begin{document}
\maketitle
\begin{abstract}
Despite the subjective nature of many NLP tasks, most NLU evaluations have focused on using the majority label with presumably high agreement as the ground truth. Less attention has been paid to the distribution of human opinions.
We collect \textbf{ChaosNLI}, a dataset with a total of 464,500 annotations to study \textbf{C}ollective \textbf{H}um\textbf{A}n \textbf{O}pinion\textbf{S} in oft-used \textbf{NLI} evaluation sets. This dataset is created by collecting 100 annotations per example for 3,113 examples in SNLI and MNLI and 1,532 examples in \abdnli.
Analysis reveals that: (1) high human disagreement exists in a noticeable amount of examples in these datasets;
(2) the state-of-the-art models lack the ability to recover the distribution over human labels; 
(3) models achieve near-perfect accuracy on the subset of data with a high level of human agreement, whereas they can barely beat a random guess on the data with low levels of human agreement, which compose most of the common errors made by state-of-the-art models on the evaluation sets. This questions the validity of improving model performance on old metrics for the low-agreement part of evaluation datasets. Hence, we argue for a detailed examination of human agreement in future data collection efforts, and evaluating model outputs against the distribution over collective human opinions.\footnote{The ChaosNLI dataset and experimental scripts are available at \url{https://github.com/easonnie/ChaosNLI}}

\end{abstract}

\section{Introduction}
Natural Language Understanding (NLU) evaluation plays a key role in benchmarking progress in natural language processing (NLP) research. With the recent advance in language representative learning~\cite{bert}, results on previous benchmarks have rapidly saturated. This leads to an explosion of difficult, diverse proposals of tasks/datasets for NLU evaluation, including Natural Language Inference (e.g., SNLI, MNLI and ANLI)~\cite{snli:emnlp2015, mnli:adina, nie2019anli}, Grounded Commonsense Inference~\cite{zellers2018swag}, Commonsense QA~\cite{talmor-etal-2019-commonsenseqa}, Social Interactions Reasoning~\cite{sap2019socialiqa}, Abductive Commonsense Reasoning (\abdnli)~\cite{bhagavatula2019abductive}, etc. 

One common practice followed by most of these recent works is to simplify the evaluation of various reasoning abilities as a classification task. This is analogous to asking objective questions to a human in educational testing. This simplification not only facilitates the data annotation but also gives interpretable evaluation results, based on which behaviors of the models are studied and then weaknesses are diagnosed~\cite{sanchez2018behavior}.

Despite the straightforwardness of this formalization, one assumption behind most prior benchmark data sourcing is that there exists a single prescriptive ground truth label for each example. The assumption might be true in human educational settings where prescriptivism is preferred over descriptivism because the goal is to test humans with well-defined knowledge or norms~\cite{trask1999key_ling}. However, it is not true for many NLP tasks due to their pragmatic nature where the meaning of the same sentence might differ depending on the context or background knowledge. 

Specifically for the NLI task, \citet{manning2006local} advocate that annotation tasks should be ``natural'' for untrained annotators, and the role of NLP should be to model the inferences that humans make in practical settings. Previous work~\cite{pavlick2019inherent} that uses a graded labeling schema on NLI, showed that there are inherent disagreements in inference tasks. All these discussions challenge the commonly used majority ``gold-label'' practice in most prior data collections and evaluations.

Intuitively, such disagreements among humans should be allowed because different annotators might have different subjective views of the world and might think differently when they encounter the same reasoning task. Thus, from a descriptive perspective, evaluating the capacity of NLP models in predicting not only individual human opinions or the majority human opinion, but also the overall distribution over human judgments provides a more representative comparison between model capabilities and `collective' human intelligence.

Therefore, we collect \textbf{ChaosNLI}, a large set of \textbf{C}ollective \textbf{H}um\textbf{A}n \textbf{O}pinion\textbf{S} for examples in several existing (English) NLI datasets, and comprehensively examine the factor of human agreement (measured by the entropy of the distribution over human annotations) on the state-of-the-art model performances.
Specifically, our contributions are:
\begin{itemize}[leftmargin=1em]
\vspace{-6pt}
\setlength\itemsep{-0.1em}
    \item We collect additional 100 annotations for over 4k examples in SNLI, MNLI-matched, and \abdnli (a total of 464,500 annotations) and show that when the number of annotations is significantly increased:
    (1) a number of original majority labels fail to present the prevailing human opinion (in 10\%, 20\%, and 31\% of the data we collected for \abdnli, SNLI, and MNLI-matched, respectively), and
    (2) large human disagreements exist and persist in a noticeable amount of examples.
    \item We compare several state-of-the-art model\footnote{We test models including BERT, RoBERTa, XLNET, ALBERT, DistilBERT, and BART.} outputs with the distribution of human judgements and show that: (1) the models lack the ability to capture the distribution of human opinions\footnote{We measure the Jensen-Shannon Distance (JSD) and the Kullback–Leibler (KL) divergence between model softmax outputs and the estimated distribution over human annotations.}; (2) such ability differs from the ability to perform well on the old accuracy metric; (3) models' performance on the subset with high levels of human agreement is substantially better than their performance on the subset with low levels of human agreement (almost close to solved versus random guess, respectively) and shared  mistakes by the state-of-the-art models are made on examples with large human disagreements.
    \item We argue for evaluating the models' ability to predict the distribution of human opinions 
    and discuss the merit of such evaluation with respect to NLU evaluation and model calibration.
    We also give design guidance on crowd-sourcing such collective annotations to facilitate future studies on relevant pragmatic tasks.
\end{itemize}
The \textbf{ChaosNLI} dataset and experimental scripts are available at \url{https://github.com/easonnie/ChaosNLI}

\section{Related Work}

\paragraph{Uncertainty of Annotations.} Past discussions of human disagreement on semantic annotation tasks were mostly focused on the uncertainty of individual annotators and the noisiness of the data collection process. These tasks include word sense disambiguation~\cite{erk2009graded, jurgens2013embracing}, coreference~\cite{versley2008vagueness}, frame corpus collection~\cite{dumitrache2019crowdsourced}, anaphora resolution~\cite{poesio2005reliability,poesio2019crowdsourced}, entity linking~\cite{reidsma2008exploiting}, tagging and parsing~\cite{plank2014learning,alonso2015learning}, and veridicality~\cite{de2012did, karttunen2014chameleon}. These works focused on studying the ambiguity of annotations, how the design of the annotation setup might affect the inter-annotator-agreement, and how to make the annotations reliable. However, we consider the disagreements and subjectivity to be an intrinsic property of the populations. Our work discusses the disagreements among a large group of individuals, and further examines the relation between the annotation disagreement and the model performance.

\paragraph{Disagreements in NLI Annotations.} Our work is significantly inspired by previous work that reveals the ``inherent disagreements in human textual inference"~\cite{pavlick2019inherent}. It employed 50 independent annotators for a ``graded" textual inference task, yielding a total of roughly 19,840 annotations, and validates that disagreements among the annotations are reproducible signals. 
In particular, in their work, the labeling schema is modified from 3-way categorical NLI to a graded one, whereas our study keeps the original 3-way labeling schema to facilitate a direct comparison between old labels and new labels, and focuses more on giving an in-depth analysis regarding the relation between the level of disagreements among humans and the state-of-the-art model performance.

\paragraph{Graded Labeling Schema.} Some previous work attempts to address the issues with human disagreements by modifying or re-defining the evaluation task with a more fine-grained ordinal or even real-value labeling schema rather than categorical labeling schema ~\cite{zhang2017joci,pavlick2019inherent,chen2019unli} to reduce the issues of ambiguity. Our work is independent and complementary to those by providing analysis on general language understanding from a collective distribution perspective.

\section{Data Collection}
Our goal is to gather annotations from multiple annotators to estimate the distribution over human opinions. Section~\ref{sec:dataset_and_task} and~\ref{sec:annotation_interface} state some details of the collection. More importantly, Section~\ref{sec:quality_control} explains the challenges of such data collection and how our designs ensure data quality.

\subsection{Dataset and Task}
\label{sec:dataset_and_task}
ChaosNLI provides 100 annotations for each example in three sets of existing NLI-related datasets. The first two sets are a subset of the SNLI development set and a subset of MNLI-matched development set, in which the examples satisfy the requirement that their majority label agrees with only three out of five individual labels collected by the original work~\cite{snli:emnlp2015, mnli:adina}.\footnote{All the examples in SNLI and MNLI development and test set come with 5 labels and the ground truth labels are defined by majority label in all previous studies. Here, we intentionally choose to label examples with a low level of human agreement in SNLI and MNLI to highlight the factor of human disagreement. Both datasets are in English.} The third set is the entire \abdnli development set introduced in~\citet{bhagavatula2019abductive}. To simplify the terminology, we denote SNLI, MNLI-m and \abdnli portion of the ChaosNLI as \chaosNLIs, \chaosNLIm, and \chaosNLIa, respectively.

\subsection{Annotation Interface}
\label{sec:annotation_interface}
To collect multiple labels for each example, we employed crowdsourced workers from Mechanical Turker with qualifications. The annotation interface is implemented using the ParlAI\footnote{\url{https://parl.ai/}}~\cite{miller2017parlai} framework. The collection is embodied in a multi-round interactive environment where at each round annotators are instructed to do one single multi-choice selection. This reduces the annotators' mental load and helps them focus on the human intelligence tasks (HITs). The compressed versions of instructions are shown in Figure~\ref{fig:instructions}. Screenshots of Turker interfaces are attached in Appendix~\ref{appendix:annotation_interface}.

\begin{figure}
    \FrameSep3pt
\begin{center}
\textbf{\footnotesize{Natural Language Inference (NLI)}}
\end{center}
\vspace{-11pt}
\begin{framed}
\scriptsize
    Given a context, a statement can be either:
\begin{itemize}[wide,labelwidth=!,labelindent=0pt,noitemsep,topsep=2pt]
    \item Definitely correct (Entailment); or 
    \item Definitely incorrect (Contradiction); or 
    \item Neither (Neutral). 
    \end{itemize}
    Your goal is to choose the correct category for a given pair of context and statement.\\
    An automatic detector will estimate your annotation accuracy on this task. 
    If your estimated accuracy is \textbf{too low}, you might be \textbf{disqualified}.\\
    If you feel \textbf{uncertain} about some examples, just choose the best category you believe the statement should be in.
\vspace{5pt}\\
\emph{Examples:\\
\textbf{Context:} A guitarist is playing in a band.\\
\textbf{Statement}: Some people are performing.\\
\textbf{Answer}: The statement is \textbf{definitely correct}.}
\end{framed}

\begin{center}
\textbf{\footnotesize{Abductive Natural Language Inference (\abdnli)}}
\end{center}
\vspace{-11pt}
\begin{framed}
\scriptsize
    Given two observations (\textbf{O-Beginning} and \textbf{O-Ending}), and two hypotheses (\textbf{H1} and \textbf{H2}), your goal is to choose one of the hypotheses that is more likely to cause \textbf{O-Beginning} to turn into \textbf{O-Ending}.\\
    An automatic detector will estimate your annotation accuracy on this task. 
    If your estimated accuracy is \textbf{too low}, you might be \textbf{disqualified}.\\
    If you feel \textbf{uncertain} about some examples, just choose the best category you believe the statement should be in.
\vspace{5pt}\\
\emph{Examples:\\
\textbf{O-Beginning:} Jenny cleaned her house and went to work, leaving the window just a crack open.\\
\textbf{H1}: A thief broke into the house by pulling open the window.\\
\textbf{H2}: Her husband went home and close the window.\\
\textbf{O-Ending}: When Jenny returned home she saw that her house was a mess.\\
\textbf{Answer}: \textbf{H1}.}
\end{framed}

\caption{Mechanical Turker instructions (compressed) for NLI and \abdnli.}
\label{fig:instructions}
\end{figure}

\subsection{Quality Control}
\label{sec:quality_control}
Collecting the ``soft-label" for examples based on plausible human opinions is difficult because we need to enforce that each annotator will genuinely try their best on the work to avoid errors caused by carelessness.
We can not denoise the data by collecting more annotations and aggregating them with majority voting, nor can we use the inter-annotator agreement to measure data quality. 

To this end, we select a set of examples , which exhibit high human agreement for a single label, to rigorously test and track the performance of each annotator. We call them \textbf{\emph{the set of unanimous examples}}.
To obtain such set, we sampled examples from SNLI, MNLI, and \abdnli training set, then crowdsourced 50 annotations for each of them, and finally selected those whose human agreement is indeed high (majority$>$95\%).
Throughout the collection process, we employ the following three mechanisms to ensure label quality:

\paragraph{On-boarding test.} Every annotator needs to pass an on-boarding test before they can work on any real example. The test includes five easy examples pre-selected from \textbf{\emph{the set of unanimous examples}}. If they fail to give the correct selection for any of them, they will be prevented from working on any example. The mechanism tests whether the annotator understands the task.

\paragraph{Training phase.} After passing the on-boarding test, each annotator will be given 10 to 12 examples from \textbf{\emph{the set of unanimous examples}} to be further annotated. For each example, if an annotator gives a label that is different from the pre-collected legitimate label, the annotator will be prompted with the correct label and told to keep concentrating on the task. If the accuracy of an annotator on training examples is below 75\%, the annotator will be disallowed to proceed. This training mechanism further helps the annotators get familiar with the task.

\paragraph{Performance tracking.} After the training phase, annotators will be given real examples. For each example to be annotated, there will be 10\% chance that the example is sampled from \textbf{\emph{the set of unanimous examples}}. Again, for such examples, if an annotator gives a label that is different from the pre-collected legitimate label, the annotator will be prompted with the correct label and told to keep concentrating on the task. If the accuracy of an annotator on those examples is below 75\% or if the annotator gives four consecutive incorrect labels, the annotator will be blocked. This mechanism tracks the performance of each annotator and guarantees that each annotator is capable and focused when working on any examples.

\begin{figure}[t!]
    \centering
    \includegraphics[width=0.98\linewidth, trim=10 15 0 10, clip]{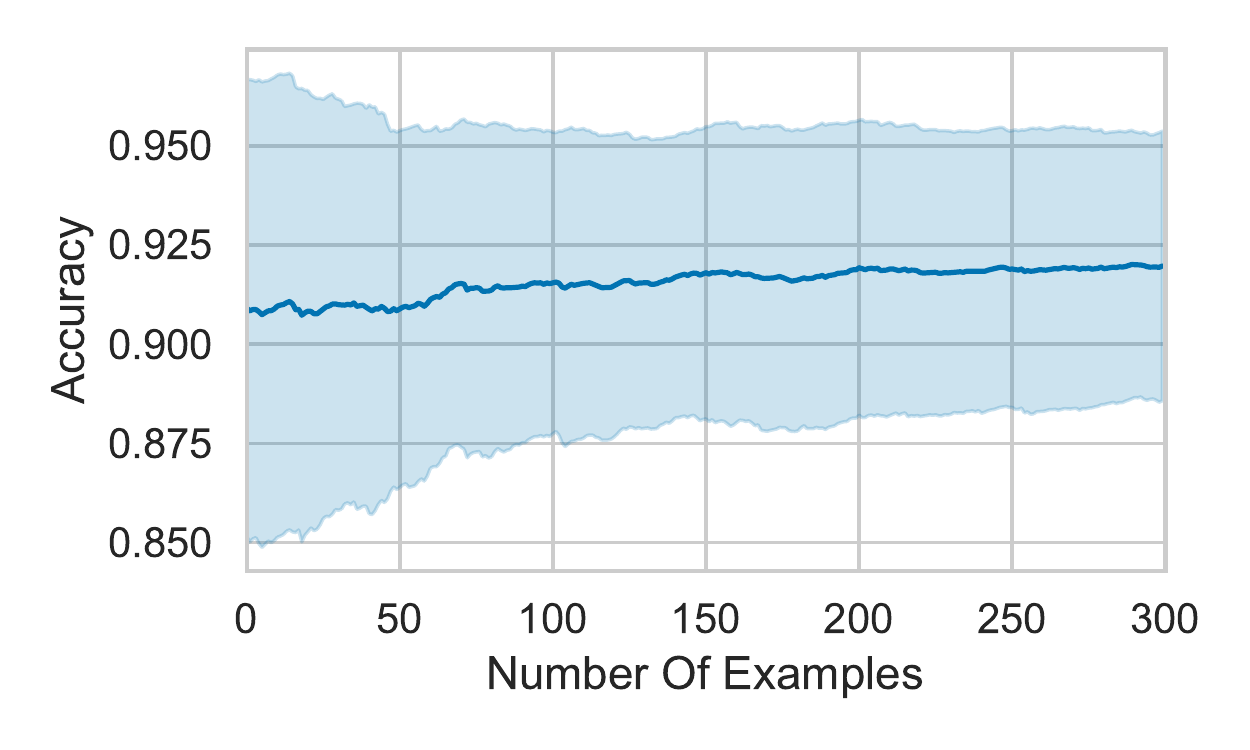}
    \caption{The accuracy range of the annotators on the NLI training and hidden \emph{unanimous examples} as they annotated their first 300 examples.}
    \label{fig:performance_trajectory}
\end{figure}

Table~\ref{tab:mturk_collection_stats} shows that on-boarding test filters more than half of the turkers.
Figure~\ref{fig:performance_trajectory} shows that the average accuracy of a single Turker on \textbf{\emph{the set of unanimous examples}} improves as the annotators have completed more examples and converges at around 92\%.\footnote{This is comparable to the accuracy of a majority voting over 5 aggregated annotations in previous work~\cite{nangia2019glue_human}.} The observations indicate that our filtering mechanisms are rigorous and help improving and keeping annotator concentrate during the collection task. The design gives guidelines for future work on how to ensure data quality where normal inter-annotator-agreement measures are not applicable.

\subsection{Other Details}
The entire collection takes about one month to complete over 464K annotations. The mean/median time a turker spent on each example ranges from 10 to 20 seconds as shown in Table~\ref{tab:mturk_collection_stats} (and we pay up to \$0.5 on average per HIT of ten examples). We observe high variance in the time/example across turkers (including over-estimation due to breaks), hence the median estimate is more reliable. We had a large set of qualified turkers for our final annotations. The total time of one month is largely attributed to the rigorous quality control mechanism via careful on-boarding qualification tests and quality monitoring.

\begin{table}[t]
    \centering
    \scalebox{0.75}{
    \begin{tabular}{lrrrr}
    \toprule
    \textbf{Data} & \textbf{QFR (\%)} & \textbf{FR (\%)} & 
    \textbf{\#Turkers} & \textbf{Time (sec)} \\
    \midrule
    \textbf{\chaosNLIa} & 7.4 & 1.3 & 1,903 & 18.7 / 12.7 \\
    
    \textbf{\chaosNLIs} & 39.9 & 14.1 & 1,639 & 15.9 / 10.1 \\
    
    \textbf{\chaosNLIm} & 39.9 & 14.1 & 1,744 & 21.2 / 13.3 \\
    \bottomrule
    \end{tabular}
    }
    \caption{MTurk statistics on the three datasets. `QFR' or Qualification Fail Rate refers to the failure rate of the onboarding qualification test. `FR' or Filter Rate refers to the ratio of Turkers who got blocked (during training phase and performance tracking described in Section~\ref{sec:quality_control}) because their performance on the \emph{unanimous examples} set are too low.
    SNLI and MNLI-m shared the same onboarding test and the same \emph{unanimous examples} set, therefore their numbers are the same. The `\#Turkers' column denotes the final set of filtered turkers that contributed to the released annotations. The last column `Time' refers to the mean  / median time spent by Turkers per example in seconds.}
    \label{tab:mturk_collection_stats}
\end{table}

\begin{table}[t]
    \centering
    \scalebox{0.90}{
    \begin{tabular}{lrrr}
    \toprule
    \textbf{Data} & \textbf{\# Examples} & \textbf{Change rate (\%)}\\
    \midrule
    \textbf{\chaosNLIa} & 1,532 & 10.64 \\
    \textbf{\chaosNLIs} & 1,514 (10k) & 24.97 (3.78)\\
    \textbf{\chaosNLIm} & 1,599 (10k) & 31.77 (5.08)\\
    \bottomrule
    \end{tabular}
    }
    \caption{Data Statistics. `\# Examples' refers to the total number of examples. `Change rate' refers to the percentage that the old majority label is different from the new majority label. The number in the parentheses shows the size of the entire original SNLI and MNLI-m development set and the percentage of label changes with respect to the entire set.}
    \label{tab:change_of_label}
\end{table}

\begin{table*}[ht]
\centering
\tiny
\begin{tabular}{p{20em}p{20em}p{6em}p{6.2em}p{1.5em}p{8em}}
\toprule
\multirow{2}{*}{\bf Context} &  \multirow{2}{*}{\bf Hypothesis} &  {\bf Old Labels} &
{\bf New Labels} & 
\multirow{2}{*}{\bf Source} & \multirow{2}{*}{\bf Type} \\
 & & \multicolumn{2}{c}{majority and individual labels} & \\
\midrule
With the sun rising, a person is gliding with a huge parachute attached to them. & The person is falling to safety with the parachute & \fulllabel{Entailment}{E E E N N} & \fulllabel{Entailment}{E\textsuperscript{(50)} N\textsuperscript{(50)}} & SNLI & Low agreements\\  
\midrule
A woman in a tan top and jeans is sitting on a bench wearing headphones. & A woman is listening to music. & \fulllabel{Entailment}{E E N N E} & \fulllabel{Neutral}{N\textsuperscript{(93)} E\textsuperscript{(7)}} & SNLI & Majority changed \\  
\midrule
A group of guys went out for a drink after work, and sitting at the bar was a real  a 6 foot blonde with a fabulous face and figure to match. & The men didn't appreciate the figure of the blonde woman sitting at the bar. & \fulllabel{Contradiction}{C N N C C} & \fulllabel{Contradiction}{C\textsuperscript{(56)} N\textsuperscript{(44)}} & MNLI & Low agreements\\  
\midrule
In the other sight he saw Adrin's hands cocking back a pair of dragon-hammered pistols. & He had spotted Adrin preparing to fire his pistols. & \fulllabel{Neutral}{N E N N E} & \fulllabel{Entailment}{E\textsuperscript{(94)} N\textsuperscript{(5)} C\textsuperscript{(1)}} & MNLI & Majority changed\\  
\bottomrule
\end{tabular}
\caption{Examples from \chaosNLIs and \chaosNLIm development set. `Old Labels' is the 5 label annotations from original dataset. `New Labels' refers to the newly collected 100 label annotations.\label{tab:nli_examples} Superscript indicates the frequency of the label.}
\end{table*}

\begin{figure*}[h]
	\centering
    \includegraphics[width=0.95\textwidth, trim=0 0 0 0, clip]{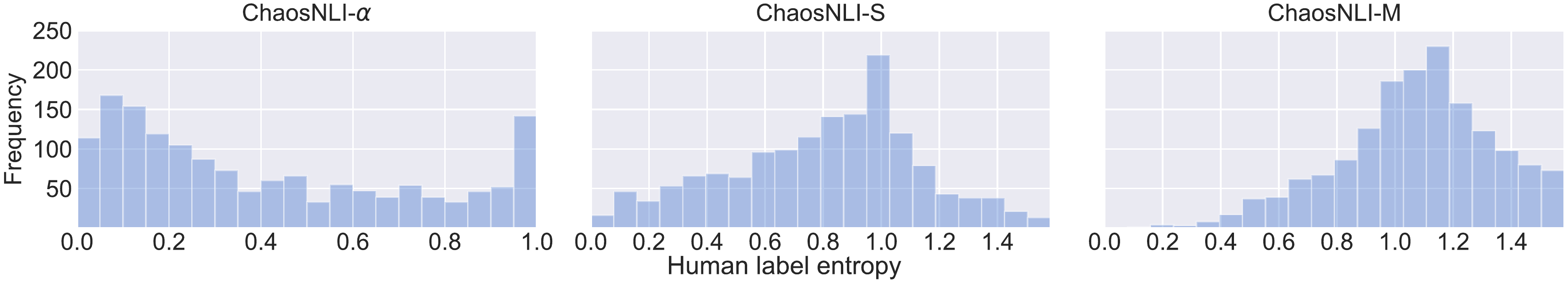}
 \caption{\label{fig:entropy_hist}Histogram of entropy of estimated distribution over human annotations on \chaosNLIa, \chaosNLIs, \chaosNLIm.}
 \vspace{-5pt}
\end{figure*}

\begin{table}[ht]
\centering
\scalebox{0.58}{
\begin{tabular}{rl}
\toprule
\textbf{Observation-1} & Sadie was on a huge hike.\\
\textbf{Observation-2} & Luckily she pushed herself and managed to reach the peak.\\
\textbf{Hypothesis-1} & Sadie almost gave down mid way.\\
\textbf{Hypothesis-2} & Sadie wanted to go to the top.\\
\textbf{Old Label} & Hypothesis-2\\
\textbf{New Labels} & \textbf{Hypothesis-1\textsuperscript{(58)}} ~~ Hypothesis-2\textsuperscript{(42)}\\
\midrule
\textbf{Observation-1} & Uncle Jock couldn't believe he was rich.\\
\textbf{Observation-2} & Jock lived the good life for a whole year, until he was poor again.\\
\textbf{Hypothesis-1} & He went to town and spent on extravagant things.\\
\textbf{Hypothesis-2} & Jock poorly managed his finances.\\
\textbf{Old Label} & Hypothesis-1\\
\textbf{New Labels} & Hypothesis-1\textsuperscript{(48)} ~~ \textbf{Hypothesis-2\textsuperscript{(52)}} \\
\bottomrule
\end{tabular}
}
\caption{Examples from the collected \chaosNLIa development set.\label{tab:abdnli_examples} The task asks which of the two hypothesis is more likely to cause \textbf{Observation-1} to turn into \textbf{Observation-2}. Superscript indicates the frequency of the label. Majority labels were marked in bold.}
\end{table}

\begin{table*}[t]
    \centering
    \scalebox{0.80}{
    \begin{tabular}{lccccccccc}
    \toprule
    \multirow{2}{*}{\textbf{Model}} 
    & \multicolumn{3}{c}{\textbf{\chaosNLIa}}
    & \multicolumn{3}{c}{\textbf{\chaosNLIs}} 
    & \multicolumn{3}{c}{\textbf{\chaosNLIm}}\\
    \cmidrule(rl){2-4}\cmidrule(rl){5-7}\cmidrule(rl){8-10}
    & \textbf{JSD$\downarrow$} & \textbf{KL$\downarrow$} & \textbf{Acc.$\uparrow$ (old/new)} & \textbf{JSD$\downarrow$} & \textbf{KL$\downarrow$} & \textbf{Acc.$\uparrow$ (old/new)} & \textbf{JSD$\downarrow$} & \textbf{KL$\downarrow$} & \textbf{Acc.$\uparrow$ (old/new)}\\
    \midrule
    \textbf{Chance} & 0.3205 & 0.406 & 0.5098/0.5052 & 0.383 & 0.5457 & 0.4472/0.5370 & 0.3023 & 0.3559 & 0.4509/0.4634\\
    \midrule
    \textbf{BERT-b} & 0.3209 & 3.7981 & 0.6527/0.6534 & 0.2345 & 0.481 & 0.7008/0.7292 & \textbf{0.3055} & \underline{0.7204} & 0.5991/0.5591\\
    \textbf{XLNet-b} & 0.2678 & \underline{1.0209} & 0.6743/0.6867 & 0.2331 & 0.5121 & 0.7114/0.7365 & \underline{0.3069} & 0.7927 & 0.6373/0.5891\\
    \textbf{RoBERTa-b} & 0.2394 & \textbf{0.8272} & 0.7154/0.7396 & 0.2294 & 0.5045 & 0.7272/0.7536 & 0.3073 & 0.7807 & 0.6391/0.5922\\
    \midrule
    \textbf{BERT-l} & 0.3055 & 3.7996 & 0.6802/0.6821 & 0.23 & 0.5017 & 0.7266/0.7384 & 0.3152 & 0.8449 & 0.6123/0.5691\\
    \textbf{XLNet-l} & 0.2282 & 1.8166 & 0.814/0.8133 & 0.2259 &  0.5054 & \underline{0.7431}/0.7807 & 0.3116 & 0.8818 & \textbf{0.6742}/\underline{0.6185}\\
    \textbf{RoBERTa-l} & \textbf{0.2128} & 1.3898 & \textbf{0.8531}/\underline{0.8368} & \underline{0.221} & 0.4937 & \textbf{0.749}/\textbf{0.7867} & 0.3112 & 0.8701 & \textbf{0.6742}/\textbf{0.6354}\\
    \midrule
    \textbf{BART} & 0.2215 & 1.5794 & 0.8185/0.814 & \textbf{0.2203} & \underline{0.4714} & 0.7424/\underline{0.7827} & 0.3165 & 0.8845 & \underline{0.6635}/0.5922\\
    \textbf{ALBERT} & \underline{0.2208} & 2.9598 & \underline{0.8440}/\textbf{0.8473} & 0.235 & 0.5342 & 0.7153/0.7814 & 0.3159 & 0.862 & 0.6485/0.5897\\
    \textbf{DistilBert} & 0.3101 & 1.0345 & 0.592/0.607 & 0.2439 & \textbf{0.4682} & 0.6711/0.7021 & 0.3133 & \textbf{0.6652} & 0.5472/0.5103\\
    \midrule
    \textbf{Est. Human} & 0.0421 & 0.0373 & 0.885/0.97 & 0.0614 & 0.0411 & 0.775/0.94 & 0.0695 & 0.0381 & 0.66/0.86\\
    \bottomrule
    \end{tabular}
    }
    \caption{Model Performances for JSD, KL, and Accuracy on majority label. $\downarrow$ indicates smaller value is better. $\uparrow$ indicates larger value is better. For each column, the best values are in bold and the second best values are underlined. ``-b'' and ``-l'' in the Model column denote ``-base'' and ``-large'', respectively.}
    \label{tab:model_performance}
\end{table*}

\section{Analysis of Human Judgements \label{sec:analysis_of_human_judgements}}

\paragraph{Statistics.}
We collected 100 new annotations for each example in the three sets described in Section~\ref{sec:dataset_and_task}. 
Table~\ref{tab:change_of_label} shows the total number of examples in the three sets and the percentage of cases where the new majority label is different from the old majority label (based on 5 annotations for SNLI and MNLI and 1 annotation for \abdnli in their original dataset, respectively). Since we only collected labels for subsets of SNLI and MNLI-m, we also include the size of the original SNLI and MNLI-m development sets and the change-of-label ratio with respect to the original sets. The findings suggest that the old labels fail to present the genuine majority labels among humans for a noticeable amount (10\%, 25\%, and 30\% for \chaosNLIa, \chaosNLIs, and \chaosNLIm, respectively) of the data. The label statistics for individual datasets can be found in Appendix~\ref{appendix:laebl_statistics}.

\paragraph{Examples.}
Table~\ref{tab:nli_examples} and Table~\ref{tab:abdnli_examples} show some collected NLI examples that either have low levels of human agreements or have different majority labels as opposed to the old ground truth labels. We can see that the resultant labels we collected not only provide more fine-grained human judgements but also give a new majority label that is better at presenting the prevailing human opinion. Moreover, there indeed exist different but plausible interpretations for the examples that are of low-level of human agreements and the discrepancy is not just noise but 
presents the distribution over human judgements with ``higher resolutions''.
This is consistent with the finding in~\citet{pavlick2019inherent}.

\paragraph{Entropy Distribution.}
To further investigate the human uncertainty in our collected labels, we show the histogram of the entropy of label distribution for \chaosNLIa, \chaosNLIs and \chaosNLIm in Figure~\ref{fig:entropy_hist}. The label distribution is approximated by the 100 collected annotations. The entropy is calculated with
$\label{equ:entropy}
\mathbf{H}\left( \bp \right) = - \sum_{i \in \mathcal{C} }{p_i \log(p_i)} 
$ and $\label{equ:human_dist}
p_i=\frac{n_i}{\sum_{j \in \mathcal{C}}n_j}$,
where $\mathcal{C}$ is the label category set and $n_i$ is the number of labels for category $i$. The entropy value gives a measure for the level of uncertainty or agreement among human judgements, where high entropy suggests low level of agreement and vice versa. 

The histogram for the \chaosNLIa shows a distribution that is similar to a U-Shaped distribution. This indicates that naturally occurring examples in \chaosNLIa are either highly certain or uncertain among human judgements. In \chaosNLIs and \chaosNLIm, the distribution shows only one apparent peak; and the distribution for \chaosNLIm is slightly skewed towards higher entropy direction.
As described in Section~\ref{sec:dataset_and_task}, \chaosNLIs and \chaosNLIm are subsets of SNLI and MNLI-m development that are of low-level of human agreements, it could be expected that the majority of naturally occurring SNLI and MNLI data would also have low entropy, which will form another peak around the beginning of the x-axis resulting a U-like shape similar to \chaosNLIa.\footnote{In our pilot study, we collected 50 labels for 100 examples of SNLI where all five original annotators agreed with each other, the average entropy of those is 0.31. The average entropy of examples on \chaosNLIs is 0.80.}

\section{Analysis of Model Predictions}
In Section~\ref{sec:analysis_of_human_judgements}, we discussed the statistics and some examples for the new annotations. The observation naturally raises two questions regarding the development of NLP models: (1) whether the state-of-the-art models are able to capture this distribution over human opinions; and (2) how the level of human agreements will affect the performance of the models. Hence, we investigate these questions in this section. Section~\ref{sec:model_and_setup} and \ref{sec:evaluation_and_metrics} state our experimental choices. Section~\ref{sec:results} discusses the results regarding the extent to which the softmax distributions produced by state-of-the-art models trained on the dataset reflects similar distributions over human annotations. Section~\ref{sec:the_factor_of_agreement} demonstrate the surprising influence of human agreements on the model performances.

\subsection{Models and Setup \label{sec:model_and_setup}}
Following the pretraining-then-finetuning trend, we focus our experiments on large-scale language pretraining models.
We studied BERT~\cite{bert}, XLNet~\cite{yang2019xlnet}, and RoBERTa~\cite{liu2019roberta} since they are considered to be the state-of-the-art models for learning textual representations and have been used for a variety of downstream tasks. We experimented on both the base and the large versions of these models, in order to analyze the parameter size factor. Additionally, we include BART~\cite{lewis2019bart}, ALBERT~\cite{lan2019albert}, and DistilBERT~\cite{sanh2019distilbert} in the experiments.
ALBERT is designed to reduce parameters of BERT by cross-layer parameter sharing and decomposing embedding. DistilBERT aims to compress BERT with knowledge distillation. BART is a denoising autoencoder for pretraining seq-to-seq models.

For NLI, we trained the models on a combined training set of SNLI and MNLI which contains over 900k NLI pairs. We used the best hyper-parameters chosen by their original authors. For \abdnli, we trained the models on \abdnli training set (169,654 examples). The hyper-parameters for \abdnli were tuned with results on \abdnli development set. Details of the hyper-parameters are in Appendix~\ref{appendix:hyperparameters}.

\subsection{Evaluation and Metrics \label{sec:evaluation_and_metrics}}
As formulated in Equation~\ref{equ:human_dist}, we used the 100 collected annotations for each example to approximate the human label distributions for each example. In order to examine to what extent the current models are capable of capturing the collective human opinions, we compared the human label distributions with the softmax outputs of the neural networks following \citet{pavlick2019inherent}.

We used Jensen-Shannon Distance (JSD) as the primary measure of the distance between the softmax multinomial distribution of the models and the distributions over human labels because JSD is a metric function based on a mathematical definition \cite{endres2003new_metric}. It's symmetric and bounded with the range [0, 1], whereas the Kullback–Leibler (KL) divergence~\cite{kullback1951information, kullback1997information} does not have these two properties.
We also used KL as a complementary measure.
The two metrics are calculated as:
\begin{equation}
    \kl{\bp \| \bq}= \sum_{i \in \mathcal{C}}p_i\log \left( \frac{p_i}{q_i} \right)
\end{equation}
\vspace{-13pt}
\begin{equation}
    \jsd{\bp\| \bq} = \sqrt{\frac{1}{2}(\kl{\bp \| \bm} + \kl{\bq \| \bm})}
\end{equation}
where $\bp$ is the estimated human distribution, $\bq$ is model softmax outputs, and $\bm = \frac{1}{2}(\bp + \bq)$.

\subsection{Main Results \label{sec:results}}
\label{sec:distance_results}
Table~\ref{tab:model_performance} reports the main results regarding the distance between model softmax distribution and estimated human label distribution. In addition to the models, we also show the results for the chance baseline (the first row) and the results for estimated human performance (the last row). 
The chance baseline gives each label equal probability when calculating the JSD and KL measures. The accuracy of the chance baseline directly shows the proportion of the examples with the majority label in a specific evaluation set. To estimate the human performance, we employed a new set of annotators to collect another 100 labels for a set of randomly sampled 200 examples on \chaosNLIa, \chaosNLIs and \chaosNLIm, respectively. For a better estimation of `collective' human performance, we ensure that the new set of annotators employed for estimating human performance is disjoint from the set of annotators employed for the normal label collection.\footnote{The estimation of collective human performance can also be viewed as calculating the JSD and KL between two disjoint sets of 100 human opinions.}
In what follows, we discuss the results.

\begin{figure*}[t]
\begin{subfigure}{0.98\textwidth}
  \centering
  \includegraphics[width=1\linewidth, trim=0 8 5 8, clip]{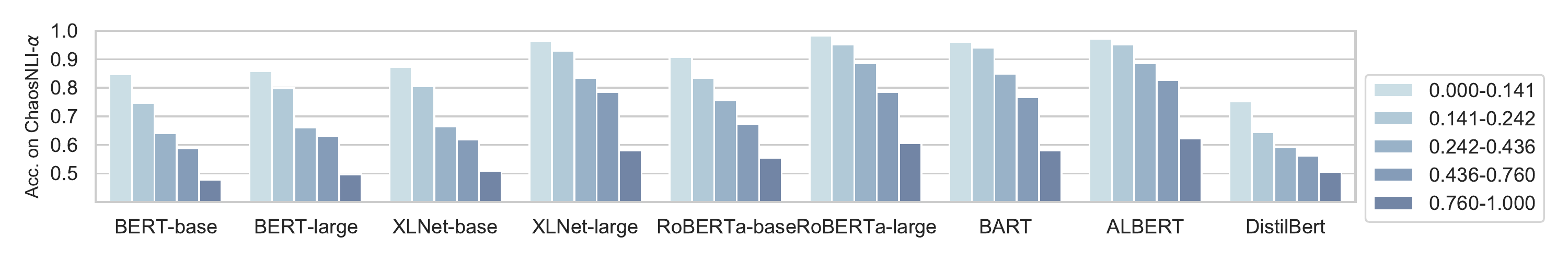}
  \vspace{-15pt}
  \label{fig:sub-test_first_bin1}
\end{subfigure}
\newline
\begin{subfigure}{0.98\textwidth}
  \centering
  \includegraphics[width=1\linewidth, trim=0 8 5 8, clip]{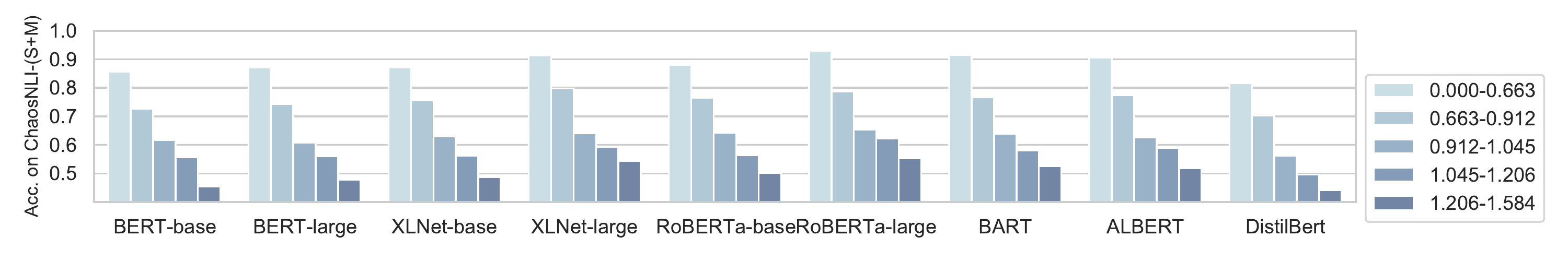}  
    \vspace{-15pt}
  \label{fig:sub-test_first_bin2}
\end{subfigure}
\caption{Accuracy on different bins of data points whose entropy values are within specific quantile ranges.}
\label{fig:bined_results}
\end{figure*}

\begin{figure*}[t]
\begin{subfigure}{0.98\textwidth}
  \centering
  \includegraphics[width=1\linewidth, trim=0 8 5 8, clip]{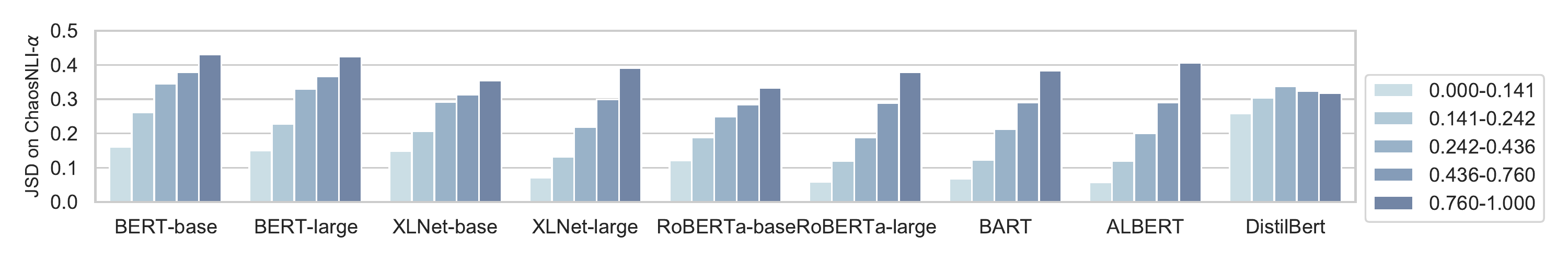}
  \vspace{-15pt}
  \label{fig:sub-test_first_jsd1}
\end{subfigure}
\newline
\begin{subfigure}{0.98\textwidth}
  \centering
  \includegraphics[width=1\linewidth, trim=0 8 5 8, clip]{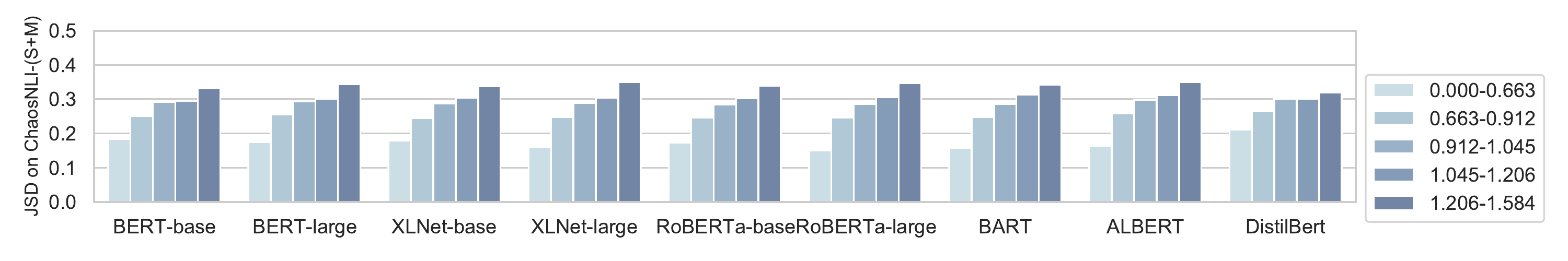}  
    \vspace{-15pt}
  \label{fig:sub-test_first_jsd2}
\end{subfigure}
\caption{JSD on different bins of data points whose entropy values are within specific quantile ranges.}
\label{fig:bined_results_jsd}
\end{figure*}

\paragraph{Significant difference exists between model outputs and human opinions.}
The most salient information we can get is that there are large gaps between model outputs and human opinions. To be specific, the estimated collective human performance gives JSD and KL scores far below 0.1 on all three sets. However, the best JSD achieved by the models is larger than 0.2 and the best KL achieved by the models barely goes below 0.5 across the table. The finding can be somewhat foreseeable since none of the models are designed to capture collective human opinions and suggests room for improvement.

\paragraph{Even chance baseline is hard to beat.}
What is more surprising is that a number of these state-of-the-art models can barely outperform and sometimes even perform worse than the chance baseline w.r.t. JSD and KL scores. On \chaosNLIm, all the models yield similar JSD scores to the chance baseline and are beaten by it on KL. On \chaosNLIa, BERT-base performs worse than the chance baseline on JSD and the scores of KL by all the models are way higher than that of the chance baseline. This hints that capturing human label distribution is a common \textit{blind spot} for many models.

\paragraph{There is no apparent correlation between the accuracy and the two divergence scores.} On both \chaosNLIs and \chaosNLIm, DistilBERT gives the best KL scores despite the fact that it obtains the lowest accuracy on the majority label. BERT-base gives the best JSD while having the second lowest accuracy on \chaosNLIm. RoBERTa-large gives the highest accuracy on \chaosNLIs and \chaosNLIm, and the second highest accuracy on \chaosNLIa but it only obtains the lowest JSD on \chaosNLIa. The best JSD score on \chaosNLIa is achieved by BART but it fails to give the best accuracy. This hints that the ability required to model the distribution of human labels differs from that required to predict the majority label and perform well on the accuracy metric.

\paragraph{Large models are not always better.}
Direct comparison between base and large models for BERT, XLNet, and RoBERTa reveals that large models cannot beat base models on \chaosNLIa and \chaosNLIm on KL scores. Moreover, on \chaosNLIm, all the large models give higher JSD scores than the base models. However, all the large models achieve higher accuracy than their base model counterparts on all three evaluation sets. This observation suggests that modeling the collective human opinions might require more thoughtful designs instead of merely increasing model parameter size.

\subsection{The Effect of Agreement \label{sec:the_factor_of_agreement}}
To study how human agreements will influence the model performance, we compute the entropy of the human label distribution (by Equation~\ref{equ:entropy}) for each data point. Then, we partition \chaosNLIa and the union of \chaosNLIs and \chaosNLIm using their respective entropy quantiles as the cut points. This results in several bins with roughly equal numbers of data points whose entropy lies in a specific range. Figure~\ref{fig:bined_results} and~\ref{fig:bined_results_jsd}  shows the accuracy and the JSD of the models on different bins.\footnote{Model JSD performances are similar to the accuracy performances where all the models obtain worse results at the bins with higher entropy range. One exception is the JSD of DistilBert on \chaosNLIa. This might due to the fact that DistilBert is highly uncertain in its prediction and tend to give even distribution for each label yielding similar results to the chance baseline.}
We observe that:
\begin{itemize}[leftmargin=1em]
\vspace{-6pt}
\setlength\itemsep{-0.2em}
    \item Across the board, \textit{there are consistent correlations between the level of human agreements and the accuracy of the model}. This correlation is positive, meaning that all models perform well on examples with a high level of human agreements while struggle with examples having a low level of human agreements. Similar trends also exists in JSD.
    \item Accuracy downgrades dramatically (from 0.9 to 0.5) as the level of human agreements decrease.
    \item The model barely outperforms and sometimes even under-performs the chance baseline on bins with the lowest level of human agreements. For both \abdnli and NLI, the accuracy of most models on the bin with the lowest level of human agreements does not surpass 60\%.
\end{itemize}
These results reveal that most of the data (which often compose the majority of the evaluation set) with a high level of human agreement have been solved by state-of-the-art models, and most of the common errors on popular benchmarks (like \abdnli, SNLI, and MNLI) lie in the subsets where human agreement is low. However, because of the low human agreement, the model prediction will be nothing more than a random guess of the majority opinion. This raises an important concern that whether improving or comparing the performance on this last missing part of the benchmarks is advisable or useful.

\section{Discussion \& Conclusion}
While common practice in natural language evaluation compares the model prediction to the majority label, Section~\ref{sec:the_factor_of_agreement} questions the value of continuing such evaluation on current benchmarks as most of the unsolved examples are of low human agreement.
To address this concern, we suggest NLP models be evaluated against the collective human opinion distribution rather than one opinion aggregated from a set of opinions, especially on tasks which take a descriptivist approach\footnote{Descriptivism is the prevailing trend in the current community compared to prescriptivism~\cite{pavlick2019inherent}.} to language and meaning, including NLI and common sense reasoning.
This not only complements prior evaluations by helping researchers understand whether model performance on a specific data point is reliable based on its human agreement, but also makes it possible to evaluate models' ability to capture the whole picture of human opinions.
Section~\ref{sec:distance_results} shows that such ability is missing from current models and potential room for improvement is huge.

It is also important to note that the level of human agreement is an intrinsic property of a data point. Section~\ref{sec:the_factor_of_agreement} demonstrates that such a property can be an indicator of the difficulty of the modeling. This hints at the connections between human agreements and uncertainty estimation or calibration~\cite{guo2017calibration} where machine learning models are required to produce the confidence value of their predictions, leading to important benefits in real-world applications.

In conclusion, we hope our data and analysis inspire future directions such as explicit modeling of collective human opinions; providing theoretical supports for the connection between human disagreement and the difficulty of acquiring language understanding in general; exploring potential usage of these human agreements; and studying the source of the human disagreements and its relations to different linguistic phenomena.

\section*{Acknowledgments}
We thank the reviewers for their helpful feedback and the authors of SNLI, MNLI, and \abdnli. This work was supported by ONR Grant N00014-18-1-2871, DARPA YFA17-D17AP00022,
NSF-CAREER Award 1846185, and DARPA MCS Grant
N66001-19-2-4031. The views contained in this article are those of the authors and not of the funding agency.

\bibliography{emnlp2020}

\begin{thebibliography}{36}
\expandafter\ifx\csname natexlab\endcsname\relax\def\natexlab#1{#1}\fi

\bibitem[{Alonso et~al.(2015)Alonso, Plank, Skj{\ae}rholt, and
  S{\o}gaard}]{alonso2015learning}
H{\'e}ctor~Mart{\'\i}nez Alonso, Barbara Plank, Arne Skj{\ae}rholt, and Anders
  S{\o}gaard. 2015.
\newblock Learning to parse with iaa-weighted loss.
\newblock In \emph{Proceedings of the 2015 Conference of the North American
  Chapter of the Association for Computational Linguistics: Human Language
  Technologies}, pages 1357--1361.

\bibitem[{Bhagavatula et~al.(2020)Bhagavatula, Bras, Malaviya, Sakaguchi,
  Holtzman, Rashkin, Downey, Yih, and Choi}]{bhagavatula2019abductive}
Chandra Bhagavatula, Ronan~Le Bras, Chaitanya Malaviya, Keisuke Sakaguchi, Ari
  Holtzman, Hannah Rashkin, Doug Downey, Scott Wen-tau Yih, and Yejin Choi.
  2020.
\newblock Abductive commonsense reasoning.
\newblock In \emph{ICLR}.

\bibitem[{Bowman et~al.(2015)Bowman, Angeli, Potts, and
  Manning}]{snli:emnlp2015}
Samuel~R Bowman, Gabor Angeli, Christopher Potts, and Christopher~D Manning.
  2015.
\newblock A large annotated corpus for learning natural language inference.
\newblock In \emph{Proceedings of the 2015 Conference on Empirical Methods in
  Natural Language Processing (EMNLP)}.

\bibitem[{Chen et~al.(2019)Chen, Jiang, Sakaguchi, and
  Van~Durme}]{chen2019unli}
Tongfei Chen, Zhengping Jiang, Keisuke Sakaguchi, and Benjamin Van~Durme. 2019.
\newblock Uncertain natural language inference.
\newblock \emph{ACL}.

\bibitem[{De~Marneffe et~al.(2012)De~Marneffe, Manning, and Potts}]{de2012did}
Marie-Catherine De~Marneffe, Christopher~D Manning, and Christopher Potts.
  2012.
\newblock Did it happen? the pragmatic complexity of veridicality assessment.
\newblock \emph{Computational linguistics}, 38(2):301--333.

\bibitem[{Devlin et~al.(2019)Devlin, Chang, Lee, and Toutanova}]{bert}
Jacob Devlin, Ming-Wei Chang, Kenton Lee, and Kristina Toutanova. 2019.
\newblock \href {https://www.aclweb.org/anthology/N19-1423} {{BERT}:
  Pre-training of deep bidirectional transformers for language understanding}.
\newblock In \emph{Proceedings of the 2019 Conference of the North {A}merican
  Chapter of the Association for Computational Linguistics: Human Language
  Technologies, Volume 1 (Long and Short Papers)}, pages 4171--4186,
  Minneapolis, Minnesota. Association for Computational Linguistics.

\bibitem[{Dumitrache et~al.(2019)Dumitrache, Aroyo, and
  Welty}]{dumitrache2019crowdsourced}
Anca Dumitrache, Lora Aroyo, and Chris Welty. 2019.
\newblock A crowdsourced frame disambiguation corpus with ambiguity.
\newblock In \emph{Proceedings of the 2019 Conference of the North American
  Chapter of the Association for Computational Linguistics: Human Language
  Technologies, Volume 1 (Long and Short Papers)}, pages 2164--2170.

\bibitem[{Endres and Schindelin(2003)}]{endres2003new_metric}
Dominik~Maria Endres and Johannes~E Schindelin. 2003.
\newblock A new metric for probability distributions.
\newblock \emph{IEEE Transactions on Information theory}, 49(7):1858--1860.

\bibitem[{Erk and McCarthy(2009)}]{erk2009graded}
Katrin Erk and Diana McCarthy. 2009.
\newblock Graded word sense assignment.
\newblock In \emph{Proceedings of the 2009 Conference on Empirical Methods in
  Natural Language Processing: Volume 1-Volume 1}, pages 440--449. Association
  for Computational Linguistics.

\bibitem[{Guo et~al.(2017)Guo, Pleiss, Sun, and
  Weinberger}]{guo2017calibration}
Chuan Guo, Geoff Pleiss, Yu~Sun, and Kilian~Q Weinberger. 2017.
\newblock On calibration of modern neural networks.
\newblock In \emph{Proceedings of the 34th International Conference on Machine
  Learning-Volume 70}, pages 1321--1330. JMLR. org.

\bibitem[{Jurgens(2013)}]{jurgens2013embracing}
David Jurgens. 2013.
\newblock Embracing ambiguity: A comparison of annotation methodologies for
  crowdsourcing word sense labels.
\newblock In \emph{Proceedings of the 2013 Conference of the North American
  Chapter of the Association for Computational Linguistics: Human Language
  Technologies}, pages 556--562.

\bibitem[{Karttunen et~al.(2014)Karttunen, Peters, Zaenen, and
  Condoravdi}]{karttunen2014chameleon}
Lauri Karttunen, Stanley Peters, Annie Zaenen, and Cleo Condoravdi. 2014.
\newblock The chameleon-like nature of evaluative adjectives.
\newblock \emph{Empirical issues in Syntax and Semantics}, 10:233--250.

\bibitem[{Kullback(1997)}]{kullback1997information}
Solomon Kullback. 1997.
\newblock \emph{Information theory and statistics}.
\newblock Courier Corporation.

\bibitem[{Kullback and Leibler(1951)}]{kullback1951information}
Solomon Kullback and Richard~A Leibler. 1951.
\newblock On information and sufficiency.
\newblock \emph{The annals of mathematical statistics}, 22(1):79--86.

\bibitem[{Lan et~al.(2019)Lan, Chen, Goodman, Gimpel, Sharma, and
  Soricut}]{lan2019albert}
Zhenzhong Lan, Mingda Chen, Sebastian Goodman, Kevin Gimpel, Piyush Sharma, and
  Radu Soricut. 2019.
\newblock Albert: A lite bert for self-supervised learning of language
  representations.
\newblock In \emph{International Conference on Learning Representations}.

\bibitem[{Lewis et~al.(2020)Lewis, Liu, Goyal, Ghazvininejad, Mohamed, Levy,
  Stoyanov, and Zettlemoyer}]{lewis2019bart}
Mike Lewis, Yinhan Liu, Naman Goyal, Marjan Ghazvininejad, Abdelrahman Mohamed,
  Omer Levy, Veselin Stoyanov, and Luke Zettlemoyer. 2020.
\newblock {BART}: Denoising sequence-to-sequence pre-training for natural
  language generation, translation, and comprehension.
\newblock In \emph{Proceedings of the 58th Annual Meeting of the Association
  for Computational Linguistics}.

\bibitem[{Liu et~al.(2019)Liu, Ott, Goyal, Du, Joshi, Chen, Levy, Lewis,
  Zettlemoyer, and Stoyanov}]{liu2019roberta}
Yinhan Liu, Myle Ott, Naman Goyal, Jingfei Du, Mandar Joshi, Danqi Chen, Omer
  Levy, Mike Lewis, Luke Zettlemoyer, and Veselin Stoyanov. 2019.
\newblock Roberta: A robustly optimized bert pretraining approach.
\newblock \emph{arXiv preprint arXiv:1907.11692}.

\bibitem[{Manning(2006)}]{manning2006local}
Christopher~D Manning. 2006.
\newblock Local textual inference: it’s hard to circumscribe, but you know it
  when you see it--and nlp needs it.
\newblock \emph{Citeseer}.

\bibitem[{{Miller} et~al.(2017){Miller}, {Feng}, {Fisch}, {Lu}, {Batra},
  {Bordes}, {Parikh}, and {Weston}}]{miller2017parlai}
A.~H. {Miller}, W.~{Feng}, A.~{Fisch}, J.~{Lu}, D.~{Batra}, A.~{Bordes},
  D.~{Parikh}, and J.~{Weston}. 2017.
\newblock {ParlAI}: A dialog research software platform.
\newblock \emph{arXiv preprint arXiv:{1705.06476}}.

\bibitem[{Nangia and Bowman(2019)}]{nangia2019glue_human}
Nikita Nangia and Samuel~R Bowman. 2019.
\newblock Human vs. muppet: A conservative estimate of human performance on the
  glue benchmark.
\newblock In \emph{ACL}.

\bibitem[{Nie et~al.(2020)Nie, Williams, Dinan, Bansal, Weston, and
  Kiela}]{nie2019anli}
Yixin Nie, Adina Williams, Emily Dinan, Mohit Bansal, Jason Weston, and Douwe
  Kiela. 2020.
\newblock Adversarial nli: A new benchmark for natural language understanding.
\newblock \emph{ACL}.

\bibitem[{Pavlick and Kwiatkowski(2019)}]{pavlick2019inherent}
Ellie Pavlick and Tom Kwiatkowski. 2019.
\newblock Inherent disagreements in human textual inferences.
\newblock \emph{Transactions of the Association for Computational Linguistics},
  7:677--694.

\bibitem[{Plank et~al.(2014)Plank, Hovy, and S{\o}gaard}]{plank2014learning}
Barbara Plank, Dirk Hovy, and Anders S{\o}gaard. 2014.
\newblock Learning part-of-speech taggers with inter-annotator agreement loss.
\newblock In \emph{Proceedings of the 14th Conference of the European Chapter
  of the Association for Computational Linguistics}, pages 742--751.

\bibitem[{Poesio and Artstein(2005)}]{poesio2005reliability}
Massimo Poesio and Ron Artstein. 2005.
\newblock The reliability of anaphoric annotation, reconsidered: Taking
  ambiguity into account.
\newblock In \emph{Proceedings of the workshop on frontiers in corpus
  annotations ii: Pie in the sky}, pages 76--83.

\bibitem[{Poesio et~al.(2019)Poesio, Chamberlain, Paun, Yu, Uma, and
  Kruschwitz}]{poesio2019crowdsourced}
Massimo Poesio, Jon Chamberlain, Silviu Paun, Juntao Yu, Alexandra Uma, and Udo
  Kruschwitz. 2019.
\newblock A crowdsourced corpus of multiple judgments and disagreement on
  anaphoric interpretation.
\newblock In \emph{Proceedings of the 2019 Conference of the North American
  Chapter of the Association for Computational Linguistics: Human Language
  Technologies, Volume 1 (Long and Short Papers)}, pages 1778--1789.

\bibitem[{Reidsma and op~den Akker(2008)}]{reidsma2008exploiting}
Dennis Reidsma and Rieks op~den Akker. 2008.
\newblock Exploiting ‘subjective’annotations.
\newblock In \emph{Coling 2008: Proceedings of the workshop on Human Judgements
  in Computational Linguistics}, pages 8--16.

\bibitem[{Sanchez et~al.(2018)Sanchez, Mitchell, and
  Riedel}]{sanchez2018behavior}
Ivan Sanchez, Jeff Mitchell, and Sebastian Riedel. 2018.
\newblock Behavior analysis of nli models: Uncovering the influence of three
  factors on robustness.
\newblock In \emph{Proceedings of the 2018 Conference of the North American
  Chapter of the Association for Computational Linguistics: Human Language
  Technologies, Volume 1 (Long Papers)}, pages 1975--1985.

\bibitem[{Sanh et~al.(2019)Sanh, Debut, Chaumond, and
  Wolf}]{sanh2019distilbert}
Victor Sanh, Lysandre Debut, Julien Chaumond, and Thomas Wolf. 2019.
\newblock Distilbert, a distilled version of bert: smaller, faster, cheaper and
  lighter.
\newblock In \emph{Proceedings of the 5th Workshop on Energy Efficient Machine
  Learning and Cognitive Computing - NeurIPS 2019}.

\bibitem[{Sap et~al.(2019)Sap, Rashkin, Chen, LeBras, and
  Choi}]{sap2019socialiqa}
Maarten Sap, Hannah Rashkin, Derek Chen, Ronan LeBras, and Yejin Choi. 2019.
\newblock Socialiqa: Commonsense reasoning about social interactions.
\newblock In \emph{EMNLP}.

\bibitem[{Talmor et~al.(2019)Talmor, Herzig, Lourie, and
  Berant}]{talmor-etal-2019-commonsenseqa}
Alon Talmor, Jonathan Herzig, Nicholas Lourie, and Jonathan Berant. 2019.
\newblock \href {https://doi.org/10.18653/v1/N19-1421} {{C}ommonsense{QA}: A
  question answering challenge targeting commonsense knowledge}.
\newblock In \emph{Proceedings of the 2019 Conference of the North {A}merican
  Chapter of the Association for Computational Linguistics: Human Language
  Technologies, Volume 1 (Long and Short Papers)}, pages 4149--4158,
  Minneapolis, Minnesota. Association for Computational Linguistics.

\bibitem[{Trask(1999)}]{trask1999key_ling}
Robert~Lawrence Trask. 1999.
\newblock \emph{Key concepts in language and linguistics}.
\newblock Psychology Press.

\bibitem[{Versley(2008)}]{versley2008vagueness}
Yannick Versley. 2008.
\newblock Vagueness and referential ambiguity in a large-scale annotated
  corpus.
\newblock \emph{Research on Language and Computation}, 6(3-4):333--353.

\bibitem[{Williams et~al.(2018)Williams, Nangia, and Bowman}]{mnli:adina}
Adina Williams, Nikita Nangia, and Samuel Bowman. 2018.
\newblock A broad-coverage challenge corpus for sentence understanding through
  inference.
\newblock In \emph{Proceedings of the 2018 Conference of the North American
  Chapter of the Association for Computational Linguistics: Human Language
  Technologies, Volume 1 (Long Papers)}.

\bibitem[{Yang et~al.(2019)Yang, Dai, Yang, Carbonell, Salakhutdinov, and
  Le}]{yang2019xlnet}
Zhilin Yang, Zihang Dai, Yiming Yang, Jaime Carbonell, Russ~R Salakhutdinov,
  and Quoc~V Le. 2019.
\newblock Xlnet: Generalized autoregressive pretraining for language
  understanding.
\newblock In \emph{Advances in neural information processing systems}, pages
  5754--5764.

\bibitem[{Zellers et~al.(2018)Zellers, Bisk, Schwartz, and
  Choi}]{zellers2018swag}
Rowan Zellers, Yonatan Bisk, Roy Schwartz, and Yejin Choi. 2018.
\newblock Swag: A large-scale adversarial dataset for grounded commonsense
  inference.
\newblock In \emph{EMNLP}.

\bibitem[{Zhang et~al.(2017)Zhang, Rudinger, Duh, and
  Van~Durme}]{zhang2017joci}
Sheng Zhang, Rachel Rudinger, Kevin Duh, and Benjamin Van~Durme. 2017.
\newblock Ordinal common-sense inference.
\newblock \emph{Transactions of the Association for Computational Linguistics},
  5:379--395.

\end{thebibliography}
\bibliographystyle{acl_natbib}

\appendix

\section{Annotation Interface}
\label{appendix:annotation_interface}
\begin{figure*}[p]
	\centering
    \includegraphics[clip,width=0.95\textwidth]{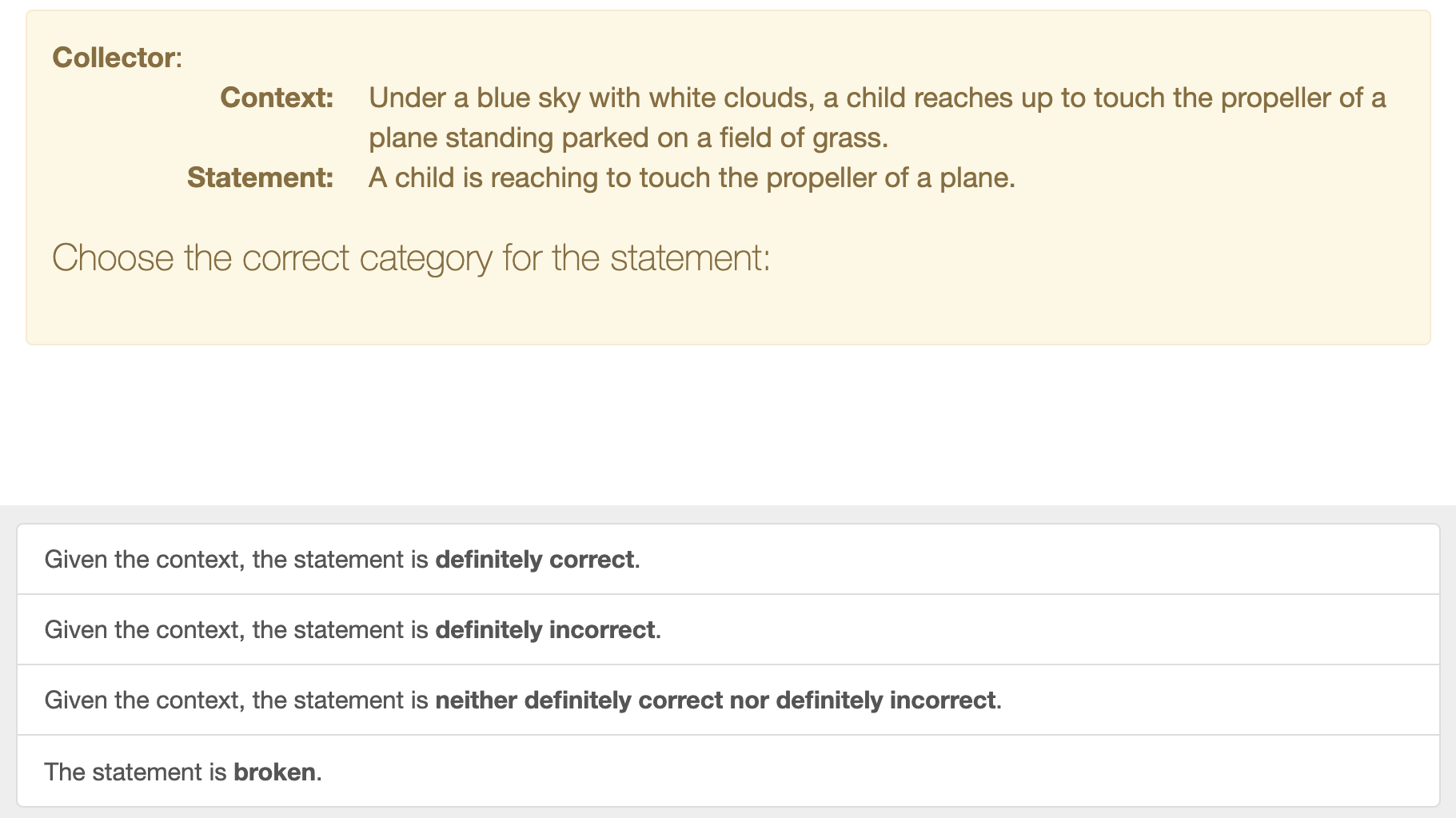}
 \caption{\label{fig:nli_interface_screenshot}Interface for NLI collection.}
\end{figure*}

\begin{figure*}[p]
	\centering
    \includegraphics[clip,width=0.95\textwidth]{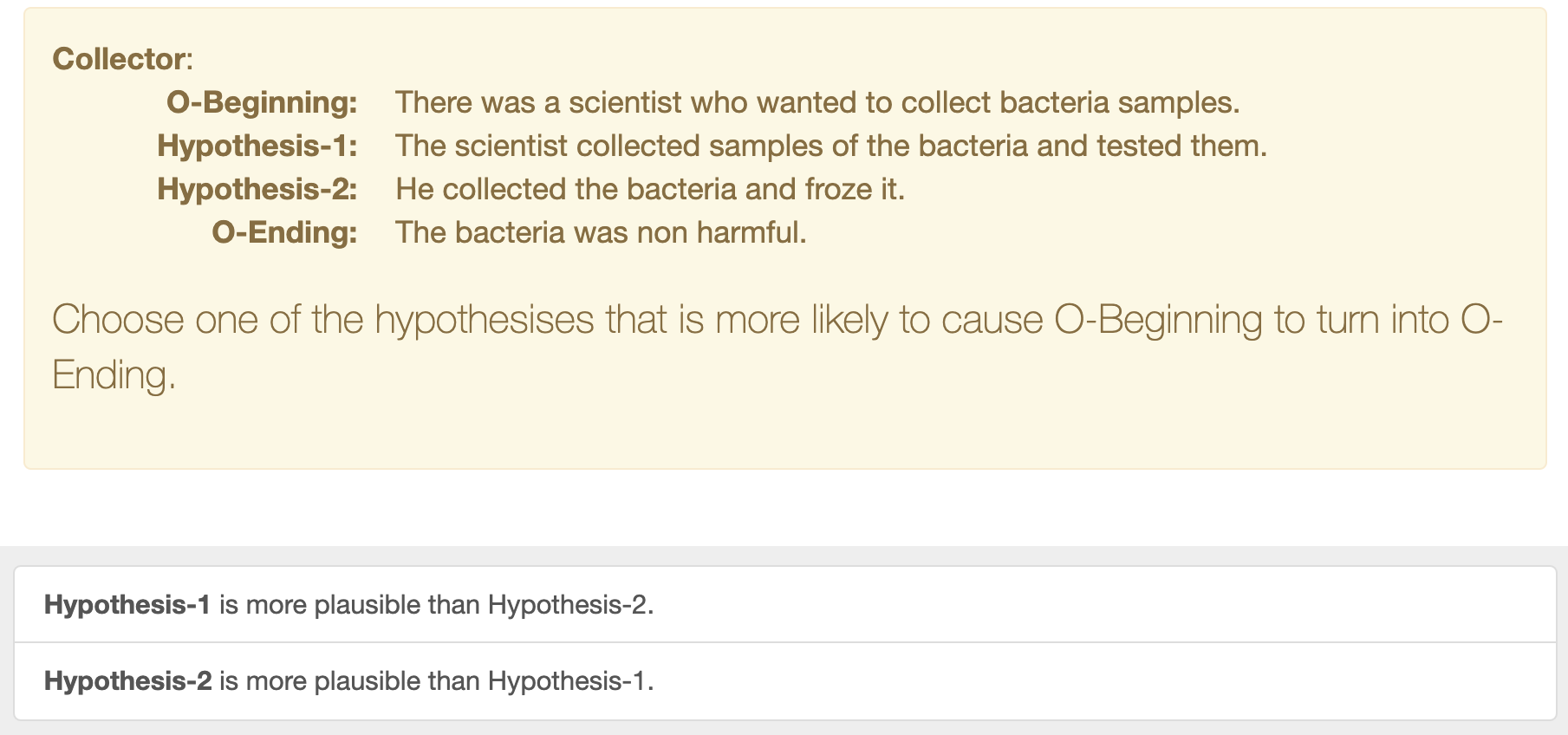}
 \caption{\label{fig:abdnli_interface_screenshot}Interface for \abdnli collection.}
\end{figure*}

Figure~\ref{fig:nli_interface_screenshot} and~\ref{fig:abdnli_interface_screenshot} show the screenshots for NLI and \abdnli collection, respectively.

\section{Hyperparameters}\label{appendix:hyperparameters}
For SNLI and MNLI, we used the same hyper-parameters chosen by their original respective authors.
For \abdnli, we tuned batch size, learning rate and the number of epoch.
For BERT, XLNet, and RoBERTa, we only searched parameters for large models and the base models use the same hyper-parameters based on the results of the large ones. Table~\ref{tab:abdnli_hyperparameters} shows the details.

\begin{figure*}[t!]
\begin{subfigure}{.24\textwidth}
  \centering
  \includegraphics[width=1\linewidth, trim=10 20 5 45, clip]{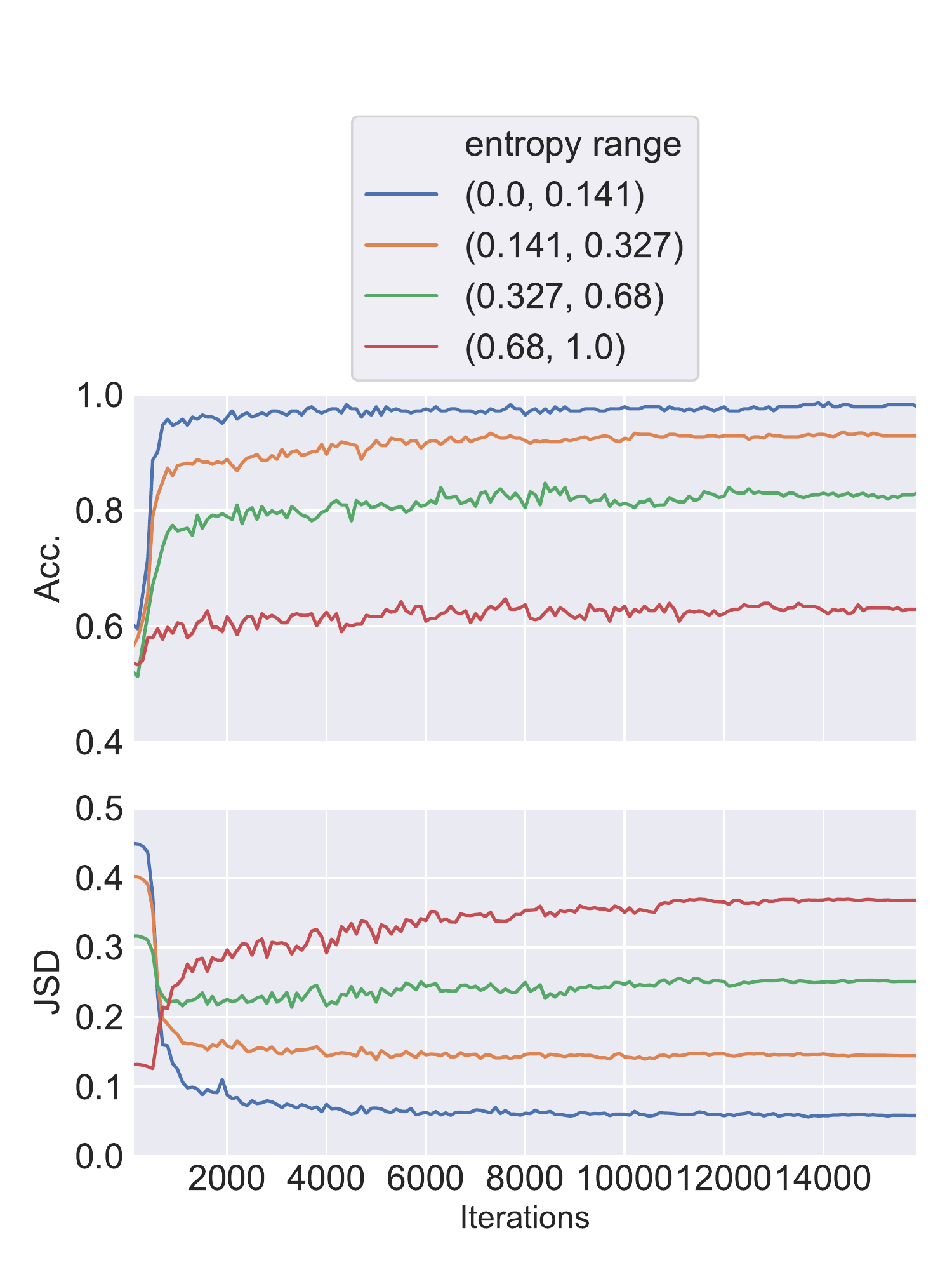}  
  \caption{ \abdnli \label{fig:sub_abdnli_trajectory}}
\end{subfigure}
\begin{subfigure}{.24\textwidth}
  \centering
  \includegraphics[width=1\linewidth, trim=10 20 5 45, clip]{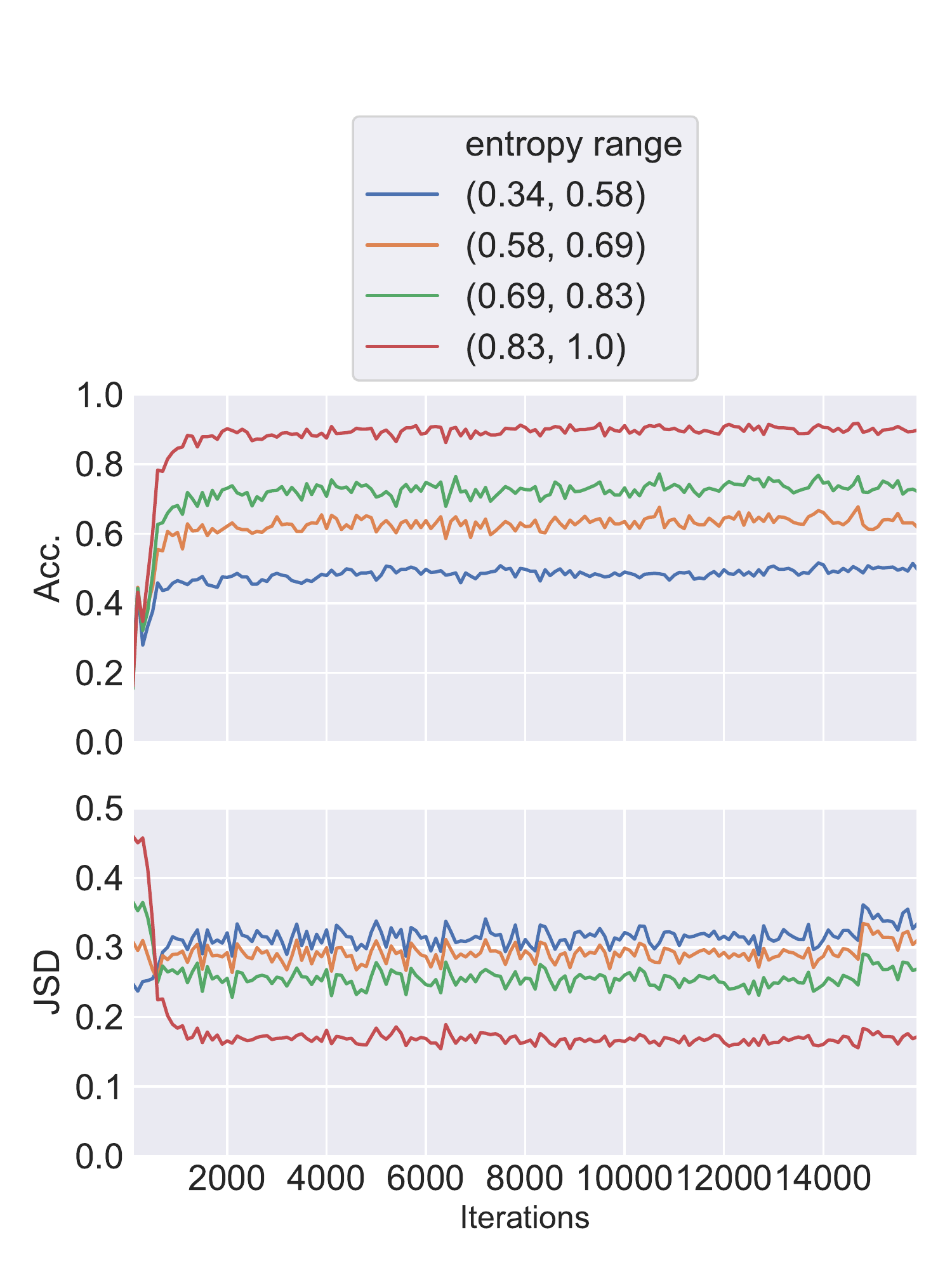}  
  \caption{{\snliat}+{\mnlimat} \label{fig:sub_nli_trajectory}}
\end{subfigure}
\begin{subfigure}{.24\textwidth}
  \centering
  \includegraphics[width=1\linewidth, trim=10 20 5 45, clip]{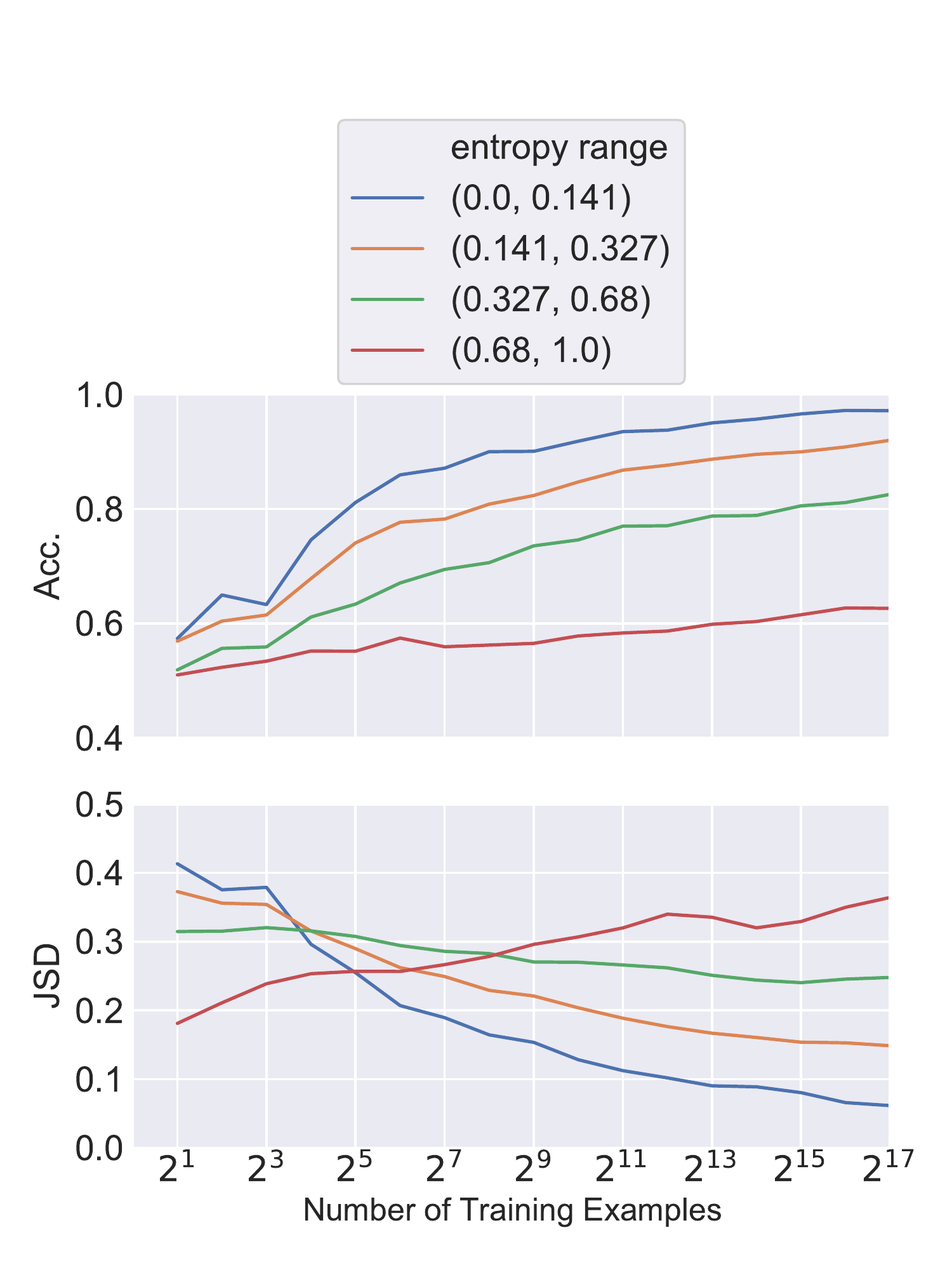}
  \caption{\abdnli}
  \label{fig:sub_abdnli_downsample}
\end{subfigure}
\begin{subfigure}{.24\textwidth}
  \centering
  \includegraphics[width=1\linewidth, trim=10 20 5 45, clip]{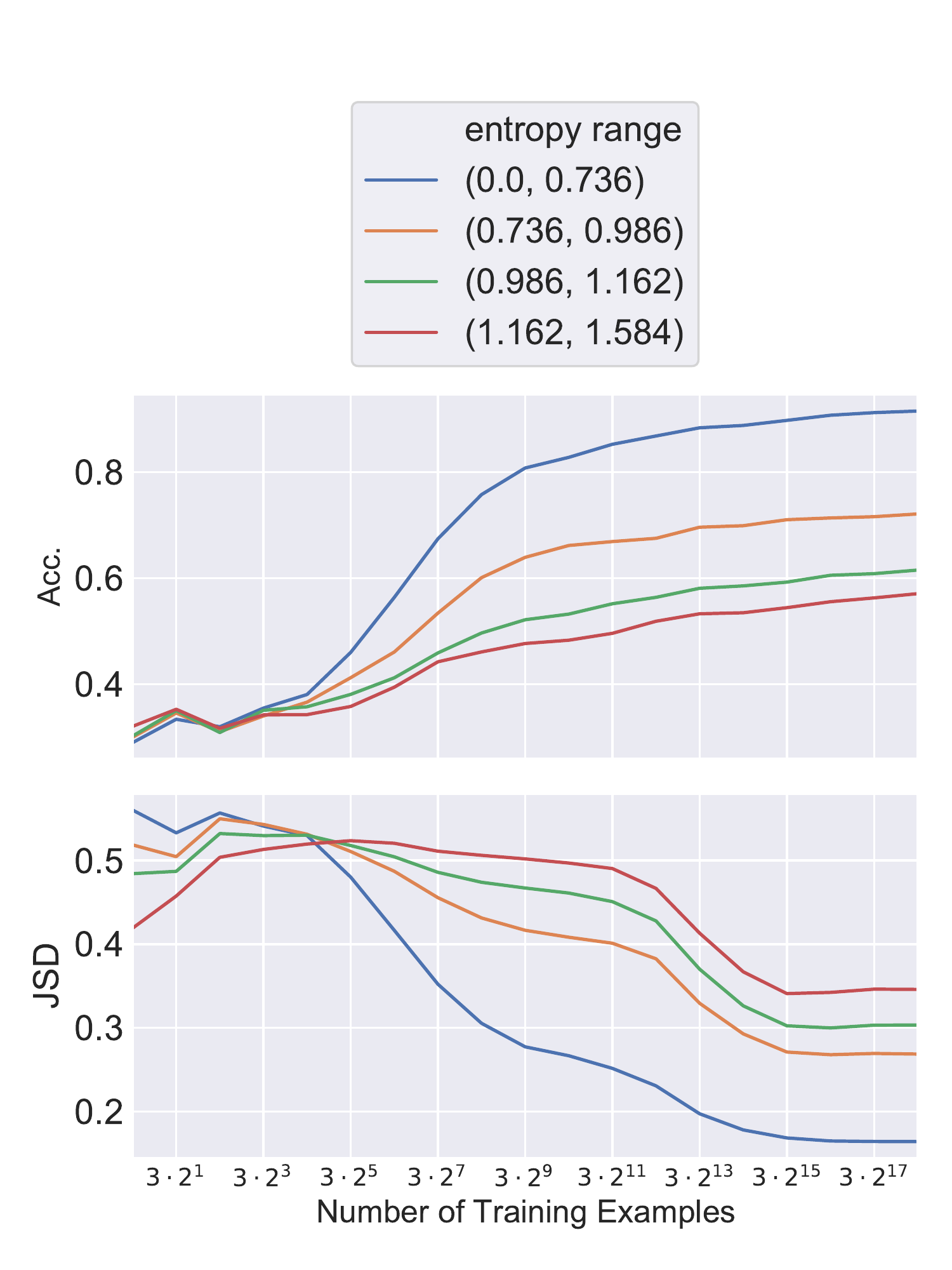}
  \caption{{\snliat}+{\mnlimat}}
  \label{fig:sub_nli_downsample}
\end{subfigure}
\caption{Sub-figure~\ref{fig:sub_abdnli_trajectory} and \ref{fig:sub_nli_trajectory} (the first two figures on the left) show the training trajectory of RoBERTa on \abdnli and {\snliat}+{\mnlimat}. Sub-figure~\ref{fig:sub_abdnli_downsample} and~\ref{fig:sub_nli_downsample} (the two on the right) show the performance curves of RoBERTa on \abdnli and {\snliat}+{\mnlimat} as the training size increased.}
\label{fig:roberta_trajectory_and_downsample}
\end{figure*}

\section{Training Size and Trajectory}\label{appendix:training_size_and_trajectory}
Figure~\ref{fig:roberta_trajectory_and_downsample} show the training trajectory and the changes of the accuracy and JSD of RoBERTa-large on four bins as the training data gradually increased in log space. The plots reveal that the accuracy of the models converges faster given fair amount of training data on bins with a high level of human agreements.

\section{Label Statistics \label{appendix:laebl_statistics}}
Labeling statistics can be found in Table~\ref{tab:nli_label_dist}. It is worth noting that there is a shift of majority labels from neutral to entailment in MNLI-m. We assume the difference might be due to multi-genre nature of the MNLI dataset, and collecting more intuitive and concrete reasons for such an observation from a cognitive or linguistic perspective will be important future work.

\section{Other Details}
Our neural models are trained using a server with a Intel(R) Xeon(R) CPU E5-2630 v4 @ 2.20GHz (10 cores) and 4 NVIDIA TITAN V GPUs. Table~\ref{tab:resource_url} shows the urls where we downloaded external resources.

\begin{table}[ht]
    \centering
    \small
    \scalebox{0.96}{
    \begin{tabular}{p{6em}p{16em}}
    \toprule
    \textbf{Resource} & \textbf{URL}\\
    \midrule
    SNLI & \url{https://nlp.stanford.edu/projects/snli} \\
    \midrule
    MNLI & \url{https://cims.nyu.edu/~sbowman/multinli}\\
    \midrule
    \abdnli & \url{http://abductivecommonsense.xyz}\\
    \midrule
    ParlAI & \url{https://parl.ai}\\
    \midrule
    Huggingface transformers & \url{https://github.com/huggingface/transformers}\\
    \bottomrule
    \end{tabular}
    }
    \caption{Links for external resources.}
    \label{tab:resource_url}
\end{table}

\begin{table*}[t]
    \centering
    \scalebox{0.80}{
    \begin{tabular}{lrrrr}
    \toprule
    \multirow{2}{*}{{ \textbf{Data}}}  & \multicolumn{3}{c}{ \textbf{Label distribution (entailment / neutral / contradiction) }} \\ \cmidrule(rl){2-5}
     & Old majority &  New majority & Old raw count (5 per ex.) & New raw count (100 per ex.)\\
    \midrule
    \textbf{\snliat} & 486 / 677 / 351 & 421\textsuperscript{(-65)} / 813\textsuperscript{(+136)} / 280\textsuperscript{(-71)} & 2,470 / 3,420 / 1,673 & 45,113 / 76,063 / 30,185\\
    \textbf{\mnlimat} & 513 / 721 / 365 & 741\textsuperscript{(+228)} / 583\textsuperscript{(-138)} / 275\textsuperscript{(-90)} & 2,483 / 3,602 / 1,910 & 64,370 / 62,794 / 32,704\\
    \bottomrule
    \end{tabular}
    }
    \caption{NLI label distribution. `Raw count' refers to the count of all individual labels. Superscript indicates the number of changes comparing to old majority labels.}
    \label{tab:nli_label_dist}
\end{table*}

\begin{table*}[ht]
    \centering
    \scalebox{0.93}{
    \begin{tabular}{lcccccc}
    \toprule
    \textbf{Hyperparam \{Search Range\}} & \textbf{BERT} & \textbf{XLNet} & \textbf{RoBERTa} & \textbf{BART} & \textbf{ALBERT} & \textbf{DistilBert} \\
    \midrule
    Learning Rate \{5e-5, 1e-5, 5e-6\} & 5e-5 & 5e-6 & 5e-6 & 5e-6 & 5e-6 & 5e-6\\
    Batch Size \{32, 64\} & 32 & 32 & 32 & 64 & 32 & 32\\
    Weight Decay  & 0.0 & 0.0 & 0.0 & 0.01 & 0.0 & 0.0\\
    Max Epochs \{3, 4, 5\} & 5 & 5 & 3 & 5 & 5 & 5\\
    Learning Rate Decay  & Linear & Linear & Linear & Linear & Linear & Linear\\
    Warmup ratio  & 0.1 & 0.1 & 0.1 & 0.1 & 0.1 & 0.1\\
    \bottomrule
    \end{tabular}
    }
    \caption{The best hyperparameters for finetuning models on \abdnli.}
    \label{tab:abdnli_hyperparameters}
\end{table*}

\end{document}